\documentclass[10pt,twocolumn,letterpaper]{article}

\usepackage{cvpr}
\usepackage{times}
\usepackage{epsfig}
\usepackage{graphicx}
\usepackage{amsmath}
\usepackage{amssymb}
\usepackage{mypac}

\usepackage[pagebackref=true,breaklinks=true,letterpaper=true,colorlinks,bookmarks=false]{hyperref}

\cvprfinalcopy 

\def\cvprPaperID{56}

\ifcvprfinal\fi
\begin{document}

\title{PatchBatch: a Batch Augmented Loss for Optical Flow}

\author{David (Dedi) Gadot ~~~~~~ Lior Wolf\\
The Blavatnik School of Computer Science\\
Tel Aviv University
}

\maketitle

\begin{abstract}
We propose a new pipeline for optical flow computation, based on Deep Learning techniques. We suggest using a Siamese CNN to independently, and in parallel, compute the descriptors of both images. The learned descriptors are then compared efficiently using the L2 norm and do not require network processing of patch pairs. The success of the method is based on an innovative loss function that computes higher moments of the loss distributions for each training batch. Combined with an Approximate Nearest Neighbor patch matching method and a flow interpolation technique, state of the art performance is obtained on the most challenging and competitive optical flow benchmarks.
\end{abstract}

\section{Introduction}

Optical flow estimation is a classical problem in computer vision. In recent works,  there has been a shift from using engineered descriptors to using Convolution Neural Networks (CNNs)~\cite{cnn} that are trained on pairs of patches that either match or do not match. Yet, the newly proposed architectures usually suffer from significant computation requirements following the tendency of using Neural Network layers as a matching function instead of traditional distance functions such as the Euclidean (L2) distance.

To support rapid computation when comparing pairs of patches, it is extremely beneficial to work in a feed-forward pipeline that encodes each patch separately and then uses conventional vector norms for comparing patches. This poses a  design restriction  on our method that is not shared with recent CNN approaches, which are optimized for accuracy at the cost of significant computation time. 

In order to achieve state of the art results despite this restriction, half a dozen novelties are brought to the field of deep patch representation. Some of the novelties arise from borrowing design choices from CNNs used for object recognition and stereo matching.

A second group of novelties are general and introduced here for the first time. We have designed new metric learning losses by augmenting the DrLIM~\cite{drlim} method. Two orthogonal augmentations are studied: the first replaces the loss of DrLIM, which is based on the potential of a spring, with a loss that is based on the potential of a centrifuge. This leads to a marked improvement on the very competitive KITTI benchmarks and in some of our synthetic experiments.  The other type of augmentation is obtained by adding, to both spring and centrifuge variants of the DrLIM loss, a term that minimizes the Standard Deviation (SD) of the two distributions: L2 distances between matching patches, and L2 distances between non-matching patches.

The two SDs are computed on the samples from each training batch and a new type of loss for training CNNs emerges. While in conventional loss functions per-sample losses are aggregated per batch, in the new type of losses the samples of the entire batch contribute jointly to the loss.

In another contribution, centered around per-batch computations, we propose a new variant of batch normalization~\cite{batchnormalization} which is more fine-grained than previously proposed. The new method improves performance but comes at a cost: the addition of these layers is not compatible with a fully convolutional deployment of the network.

\subsection{Method overview}

The optical flow solution described below is comprised of a series of well established building blocks, where at the heart of the pipeline lies a novel way to compute descriptors and compare patches. 

First, a  pair of gray-level input images is normalized by subtracting from each image its mean and dividing by its SD. We then compute, independently and in parallel, descriptors per each pixel of each image. These descriptors are learned from examples using a Siamese Deep Neural Network architecture.

The PatchMatch~\cite{patchmatch} (PM) method is then used as an Approximate Nearest Neighbor (ANN) algorithm on top of the learned descriptors. The conventional L2 metric is used, thus simplifying the ANN computation, as opposed to previous works which use Neural Network layers to compute the matching score.

We then employ a bidirectional consistency check and eliminate all non-consistent matches. In addition, we remove small independent clusters of flow predictions using a connected component analysis. 

The surviving matches provide sparse optical flow. The flow maps are downsampled, for computational reasons, by a factor of $2$ and $4$ for the KITTI datasets and the MPI-Sintel dataset respectively. The decimated maps are then given as input to the EpicFlow~\cite{epicflow} algorithm, which interpolates the correspondence fields and creates a dense optical flow map.

\section{Previous work}
\label{sec:prevwork}
Dense optical flow methods have been the subject of research for the past 35 years, starting with the work of Horn and Schunk~\cite{hornschunk}. At the beginning, optical flow research was limited to small displacements only. A significant advancement occurred with the work of Brox and Malik~\cite{broxmalik} who were the first to provide reasonable performance for large displacements.

The three major modern datasets in the field are KITTI2012~\cite{kitti2012},  which is a real world database consisting of images taken from a moving vehicle; MPI-Sintel~\cite{mpisintel}, which is a synthetic database consisting of computer-created movies; and the latest KITTI2015~\cite{kitti2015}, which is a new real world database in which both the camera and the scene are non-stationary.

It is common to distinguish between 2-frame (pure) optical flow methods and methods that require more complex inputs. While the former relies solely on an input of two sequential images, the latter may employ stereo images, more than two input images, etc. Out of the 2-frame optical flow algorithms the FlowFields method~\cite{flowfields} provides an elaborate pipeline, somewhat similar to ours, and presents near state-of-the-art performance on the KITTI2012 database. There are several significant differences between our work and~\cite{flowfields}. The most important is that the latter uses engineered features while we use CNNs in order to compute initial correspondences. Another prominent algorithm is PH-Flow~\cite{phflow}, which brings state-of-the-art performance to the KITTI2012 database, at the cost of extensive computation time. 

An additional reference work is EpicFlow~\cite{epicflow}. In the original paper the authors used a matching technique called DeepMatching~\cite{deepmatching} in order to compute a sparse correspondence field that is then being interpolated in order to create a dense flow field. The interpolation is based on edge-aware averaging of the sparse correspondence field. In our work we employ EpicFlow's interpolation technique on a sparse correspondence field which is calculated using our descriptors and PatchMatch~\cite{patchmatch}.  An alternative method for interpolating a dense optical flow, which has also been applied to DeepMatching-based inputs is DeepFlow~\cite{deepflow}.

As previously mentioned, our pipeline employs the PM algorithm~\cite{patchmatch} in order to compute the initial correspondence field. 
PM uses the inherent smoothness and coherency of natural images in order to propagate accurate ``guesses'' between neighboring pixels, in addition to a random-search stage which helps to avoid local minima. Image-based ANN alternatives include variants such as TreeCANN~\cite{treecann} and CSH~\cite{csh}. However, we chose PM as our ANN algorithm due to its simplicity, efficiency and modularity. These properties allow us to modify PM for our pipeline. When using PM, instead of utilizing image-based patches as inputs, we use our own features, which were computed using a CNN architecture as mentioned before.

Recent advancements in Deep Learning did not skip the fields of Optical Flow and Stereo Matching. In the FlowNet pipeline~\cite{flownet}, a CNN was presented that conducts almost the entire optical-flow computation inside the neural network. Though not achieving state-of-the-art results on any of the major datasets, their network runs in real-time and opens a gate to other (almost) complete end-to-end solutions being computed with a single neural network. 

In a recent work on stereo matching~\cite{stereomatching}, a CNN architecture compares two candidate stereo patches, followed by extensive post-processing. Each of the patch pairs goes through several identical computations. The resulting activations are then combined and processed through similarity computing layers. However, computational efficiency would be much higher, if an L2 distance of the separate activations would be used instead~\cite{newlecun}. This makes our architecture fully-convolutional, allowing an improved run time. An additional difference is that while~\cite{stereomatching} uses small $9\times9$ patches, we found that larger patches are beneficial, and our main architecture uses $51\times51$ patches. 

Computing patch similarity using deep networks is a thoroughly investigated subject. In~\cite{patchsimilarity} the authors inspected several CNN architectures which are able to produce a patch-similarity score. Their conclusion was that there is a sizable advantage for computing the final similarity score using a complex function that involves several dense layers (see also~\cite{stereomatching}). Yet, having to pass every two patches through a comparison network leads to a sharp increase in run time. Thus we chose a different path and insist on using per-patch representations that support L2 distance comparisons. This is done in order to reduce the method's computational complexity in the ANN computation stage. 

In order to learn patch representations that can be effectively compared using the L2 norm, we employ several variants of the DrLIM method~\cite{drlim}, which is widely used to learn similar from non-similar. However, there are only a few variants of it in the literature.

A major contribution of our work is the incorporation of per-batch statistics, collected during training. Somewhat related is the Batch Normalization method~\cite{batchnormalization}, which takes advantage of batch-based statistics in order to normalize the activations and accelerate the network's training, and avoid some of the local minima. This is different from our usage of batch statistics for augmenting the loss itself.

In addition to using batch statistics in order to incorporate distribution information to the loss, we also expand the idea of batch normalization to allow fine-grained control of the network's convergence. This is done by performing the normalization at each activation and not at the level of the entire layer as is done in~\cite{batchnormalization}. 

\section{Network architecture} 
\label{sec:networkarchitecture}

We study patches of typical sizes of $51 \times 51$ or $71 \times 71$. This is similar to the $64\times 64$ patches used in~\cite{patchsimilarity}. In addition, unlike previous work~\cite{patchsimilarity}, we do not employ patches at multiple scales in our network. While color information might be be useful, e.g., on MPI-Sintel, we discard color since the KITTI2012 benchmark is grayscale. 

We train a fully-convolutional neural network, which creates descriptors we later use in the matching process. Inspired by modern object recognition networks~\cite{vgg}, we use small $3\times 3$ filters in each convolution layer other than the last. The network is built out of a repeating pattern of three layers such that each layer triplet is a combination of a convolutional layer, a batch-normalization layer, and a max-pooling layer. In the last layer triplet we omit the max-pooling layer and use $2 \times 2$ filters. Leaky ReLU~\cite{relu}, with a parameter of $0.1$ is used as the non-linearity following each convolutional layer, including the last one. Overall we use 5 such structures, see Tab.~\ref{tab:network} within a Siamese architecture~\cite{siam}. While some may claim that max-pooling layers hinder matching accuracy by causing the network to become translation-invariant, when using our architecture one can observe no such phenomena.

\begin{table}[t!]
\begin{center}
\begin{tabular}{|l|c|c|c|c|}
\hline
Layer           & Filter/Stride &  Output size \\ \hline
Input        & -- & $1 \times 51 \times 51 $\\
\hline
Conv1        & $3 \times 3 $ / 1                   & $32\times 49 \times 49 $          \\ 
Batch Normalization   &--                   & $32\times 49 \times 49 $     \\

Max Pool        & $2 \times 2$ / 2                & $ 32 \times 25 \times 25 $                            \\ \hline
Conv2        & $3 \times 3 $ / 1                   & $64\times 23 \times 23 $          \\ 
Batch Normalization   &--                   & $64\times 23 \times 23 $     \\

Max Pool        & $2 \times 2$ / 2                & $ 64 \times 12 \times 12 $                            \\ \hline
Conv3        & $3 \times 3 $ / 1                   & $128 \times 10 \times 10 $          \\ 
Batch Normalization   &--                   & $128 \times 10 \times 10 $     \\

Max Pool        & $2 \times 2$ / 2                & $ 128 \times 5 \times 5 $                            \\ \hline
Conv4        & $3 \times 3 $ / 1                   & $256\times 3 \times 3 $          \\ 
Batch Normalization   &--                   & $256\times 3 \times 3 $     \\

Max Pool        & $2 \times 2$ / 2                & $ 256 \times 2 \times 2 $                            \\ \hline
Conv5        & $2 \times 2 $ / 1                   & $ 512\times 1 \times 1 $          \\ 
Batch Normalization   &--                   & $512\times 1 \times 1 $     \\
\hline
\end{tabular}
\end{center}
\caption{The network model for representing a grayscale $51 \times 51$ input patch as a $512D$ vector. The Batch Normalization used is our fine-grained variant. Leaky ReLU units~\cite{relu} (with $\alpha = 0.1$) are used as activation functions following the five batch normalization layers.}
\label{tab:network}
\end{table}

We employ a variant of the batch normalization layer, which differs from the conventional batch normalization method~\cite{batchnormalization}. While the latter employs a single value of mean, SD, $\gamma$, and $\beta$ parameters for each feature map, our variant computes these parameters for each single pixel. For example, for the output of the first convolutional layer, there are $32 \times 49 \times 49 \times 2$ learned parameters ($\gamma$ and $\beta$) and the same number of computed batch statistics (mean and SD, for each activation in the volume). Once computed, each activation is normalized by subtracting the mean, dividing by the SD and it then undergoes a scale and shift transformation: $y_i = \gamma \hat x_i + \beta$, where $\hat x_i$ is the normalized activation and $y_i$ is the post-transformation value.

As shown by our experiments, this modification creates a significant gap in performance, see Sec.~\ref{sec:exp}. However, it comes at a cost: the need to normalize each pixel separately does not allow for an efficient fully convolutional computation of the descriptors. Instead, it requires a much slower sliding window approach. We therefore also study an alternative architecture called FAST in which the batch normalization process is done in a conventional way.

The network is strictly-Siamese, which allows us to later compute the descriptors of each image independently. The matching cost is computed using a simple L2 metric. The patch representation (descriptor) size is typically of length 512. Experiments reveal that using a larger descriptor leads to a small increase in accuracy, and using a descriptor size as small as 32 leads to only a moderate loss of accuracy. 

\subsection{Loss}
\label{sec:loss}

Architectures similar to the one described above were explored by previous work~\cite{patchsimilarity,newlecun}. In each previous work, such per-patch architectures were found to be significantly inferior to the architectures that use two patches as inputs. Much of the improved performance we present in this work can be attributed to the novel variants of the DrLIM's loss employed, which are explored next.

The conventional DrLIM loss, which is motivated by the spring model is given by
\bea
\label{eq:spring}
(1-Y) \frac{1}{2}D_w^2 + (Y)\frac{1}{2}\{\max (0,m-D_w) \}^2~,
\eea
where, $Y=0$ for matching pairs, $Y=1$ otherwise, $m$ is the margin parameter, and $D_w$ is the L2 distance between the pair of samples.

We suggest two orthogonal modifications. The first modification is to insert the square into the hinge and obtain the following formula: 
\bea
\label{eq:cent}
(1-Y) \frac{1}{2}D_w^2 + (Y)\frac{1}{2}\{\max (0,m^2-D_w^2) \}~.
\eea

Whereas the original DrLIM was motivated by the spring model analogy~\cite{drlim}, the new loss can be said to model a sticky centrifuge. Let $M$ be a mass of a particle located at rest at a distance $r$ in a frame rotating at an angular velocity $\omega$ around the origin. The particle feels the centrifugal force $\vec{F} = m\omega^2 r\hat{r}$ in direction $\hat{r}$. This force is derived from the potential $V\left( r \right) =  - M{\omega ^2}{r^2}$ as $\vec{F} = - \nabla V$. Assuming that the centrifuge has a sticky boundary at a radius $m$, particles at a radius larger than $m$ would just rotate with the centrifuge. The potential then becomes $V_{cen}\left( r \right) =M{\omega ^2}\max \left( {0,{m^2} - {r^2}} \right)$.

\begin{figure*}[!t]
\centering
\begin{tabular}{ccc}
\includegraphics[width=.31\linewidth]
{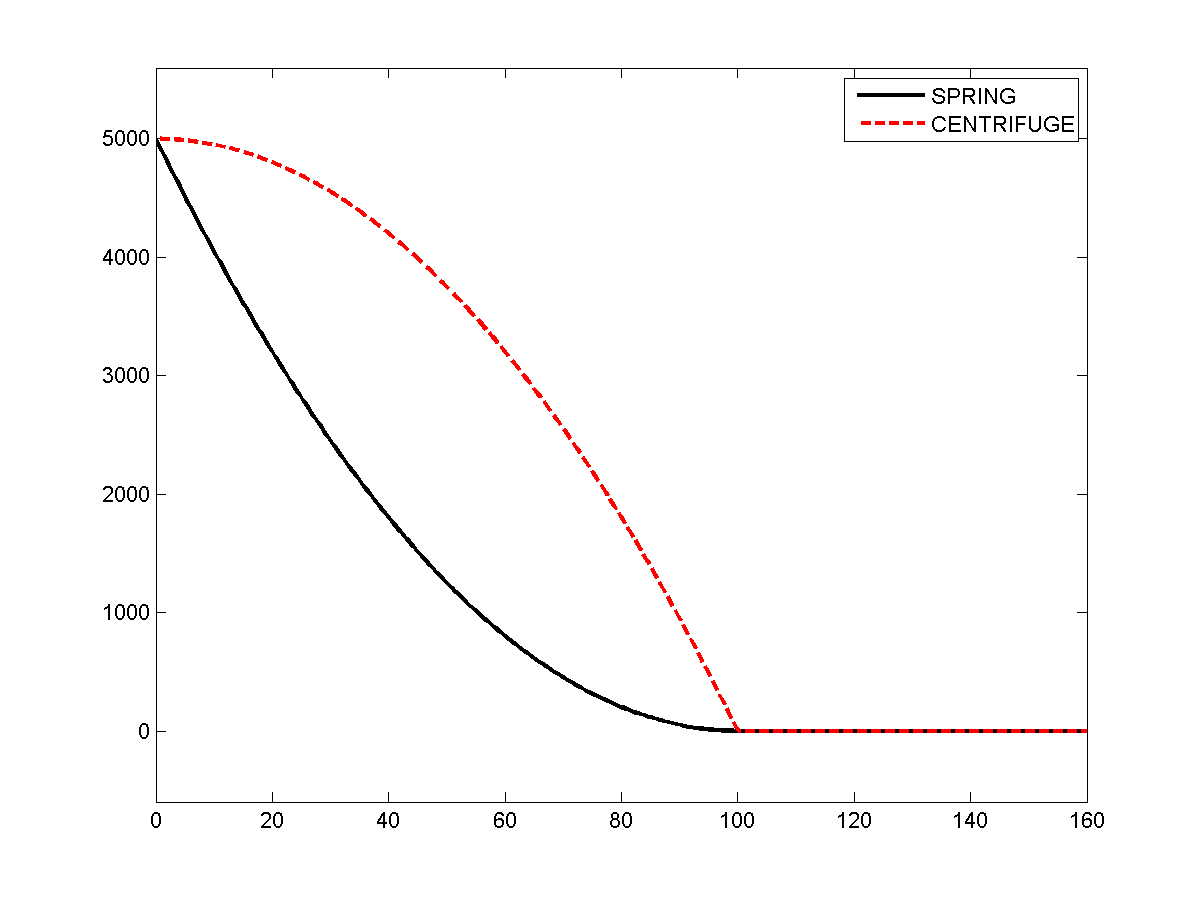}&
\includegraphics[width=.31\linewidth]{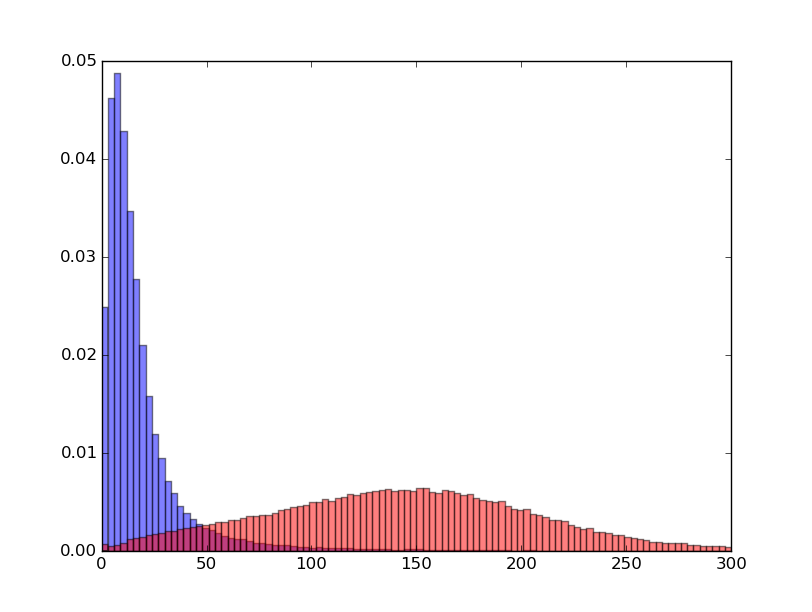}&
\includegraphics[width=.31\linewidth]{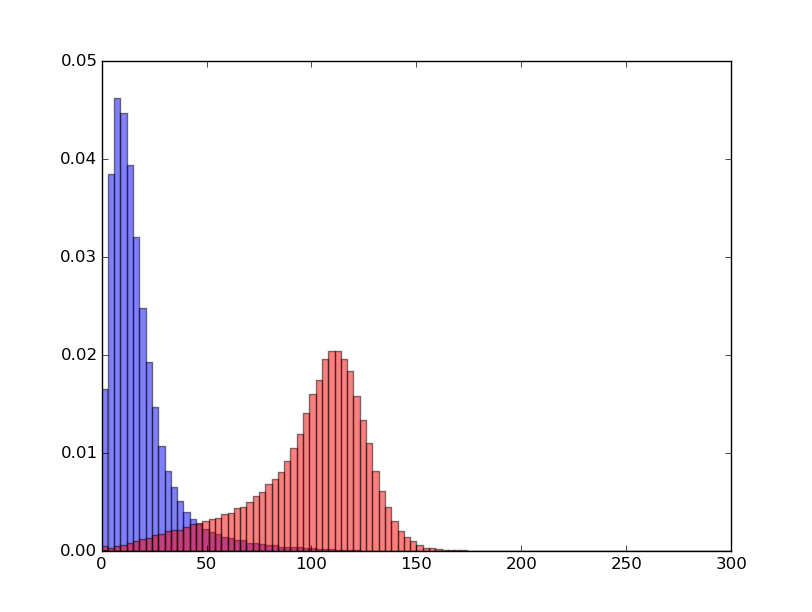}\\
(a) & (b) & (c)\\
\end{tabular}
\caption{Demonstrating the effect of DrLIM variants. (a) A comparison of the loss (y-axis) on negative pairs as a function of the distance $D_w$ (x-axis) for $m=100$ for the original DrLIM (SPRING) and the CENTRIFUGE variant. The CENTRIFUGE is more sensitive to value shifts near the margin and then loses its sensitivity. (b) The distribution of distances for matching (left side, blue) and non-matching (right side, red) pairs, for KITTI2012 validation data, when using the CENTRIFUGRE variant. The plot shows distance vs. frequency. (c) The same two distributions for the CENTRIFUGE+SD loss. Adding the batch SD to the loss causes the means to be somewhat closer. However, the SD of both distributions is much reduced.}
\label{fig:losses}
\end{figure*}

Based on the underlying physical models, the terms SPRING and CENTRIFUGE will be used below to refer to the conventional DrLIM of Eq.~\ref{eq:spring} and the variant of Eq.~\ref{eq:cent}. Fig.~\ref{fig:losses}(a) depicts the shape of the loss functions on the negative (Y=1) pairs.

The second modification we add to the DrLIM loss is based on per-batch statistics. The augmented loss then incorporates these statistics, unlike any loss in the literature we are aware of. The effect of this modification can be dramatic, as can be observed in Fig.~\ref{fig:losses}(b),(c).

The batch statistics we consider are the SD of the distances of the two classes -- matching and non-matching. The basic motivation for this strategy is the need to increase the separation between the two distributions. While the DrLIM loss pulls the samples to be close to either $0$ or $m$, we found the two distributions to overlap considerably. Adding the requirement of a small SD directly pulls the two distributions closer to their respective means and improves separability.

Let $\sigma_Y$, $Y=0,1$ be the SD value, in a training batch, of the pairwise distance $D_w$ for samples that match or do not match, respectively. The SPRING+SD variant is defined as:
\begin{multline}
\label{eq:spring+sd}
(1-Y) \lambda D_w^2 + (Y) \lambda \{\max (0,m-D_w) \}^2 + (1-\lambda) (\sigma_0+\sigma_1)
\end{multline}
\noindent The CENTRIFUGE+SD variant is given by:
\begin{multline}
\label{eq:cent+sd}
(1-Y) \lambda D_w^2 + (Y)\lambda \{\max (0,m^2-D_w^2) \}+(1-\lambda) (\sigma_0+\sigma_1)
\end{multline}

In both variants, a parameter $\lambda$ is added which controls the tradeoff between the core DrLIM variants and the augmentation by the standard deviation. In all the experiments in this paper $\lambda$ was set to a value of $0.8$.

\subsection{Training}
Each image of a chosen dataset is normalized by subtracting its own mean and dividing by its own SD. The same normalization is later used during test time. We sample two populations, matching and non-matching, by collecting $51\times51$ patches and using the given ground-truth flow computation. For the non-matching population we employ a random shift from the ground truth in both the X and Y axes of 1-8 pixels. Requiring even small translations to become non-matching is in contrast, for example, with~\cite{stereomatching}, which used 4-8 pixels for the non-matching class. 

In order to augment the data, flips and 90 degree rotations are applied on-the-fly during train time. We use AdaDelta~\cite{adadelta} as an efficient, adaptive, learning rule and Lasagne~\cite{lasagne}, which is a Theano~\cite{theano} based Deep Neural Network framework. 
We trained the final network for 4000 epochs, in each epoch we used 50,000 random samples from our created database with a batch-size of 256.

\section{Matching and Interpolation}
Since our architecture is strictly Siamese, we can compute the features of each image independently and in parallel. Calculating descriptors using the FAST architecture takes approximately 2 seconds per image. Using the ACCURATE architecture is more time consuming, due to the fact that the image is being split to patches and each patch descriptor is then computed independently in a sliding window manner. This takes approximately 27 seconds per image using an NVIDIA Titan X GPU.

\subsection{Matching}
\label{sec:pm}
When using PM as an ANN algorithm, we use as input our created descriptors and not the gray-scale image patches, similar to the Generalized PM~\cite{gpm} approach. The squared L2 distance is used as the matching metric. 

We only run PM for two iterations in order to reduce the computation time and also, more importantly, since it was found that adding iterations causes additional matching outliers to appear. The same phenomenon was described in~\cite{flowfields}: the additional iterations of the ANN used there were said to create ``resistant outliers'', whose matching distances are below those of the true matches. 

PM is used twice, in parallel, from the first image to the second and vice-versa, in order to check for the consistency of the two flow fields. All matches which do not exactly point to one another in this bidirectional consistency check are being eliminated (PM's output is an integer assignment). 

It was found empirically that allowing a large random-search radius during the PM process helped to improve performance on the KITTI datasets while we saw no such effect on the MPI-Sintel dataset. This observation is consistent with the average highest disparity for each image-pair in the different datasets. Following these observations the random search parameter of PM was set to 500 on the KITTI datasets, and to only 10 on the MPI-Sintel dataset.

Following the bidirectional consistency check, a binary mask indicating reliable flows is considered, and its connected components are identified. Small connected components are then considered unreliable. Specifically, we use a threshold area of 10,000 for the KITTI datasets and 400 for MPI-Sintel. For the MPI-Sintel dataset we also eliminate all the matches around the borders of the image (30 pixels) since we have found that there are more outliers there than in the rest of the image, probably due to the relatively large patch size we are using.

\subsection{Interpolation}
\label{sec:interp}

Given a sparse correspondence field, describing the matches which met the bidirectional consistency criterion and the connected component filtering, we employ EpicFlow~\cite{epicflow} to obtain a dense correspondence field. The EpicFlow algorithm interpolates each missing prediction using its neighboring predictions from the sparse correspondence field, i.e. its support. From this support, a number of affine transformations are calculated using multiple subsets of correspondences. An edge map is computed using the SED method~\cite{sed}, and the affine transformations are then averaged based on the geodesic distance computed from the image's edge map.

\section{Experiments}


In order to demonstrate the effectiveness of the new DrLIM variants beyond the scope of optical flow computations, we have conducted a series of synthetic experiments in addition to testing the impact of the new variants on real datasets. 

In the first experiment, $n_c$ multivariate Gaussian centers are uniformly sampled from a $256D$ hypercube of edge length $1$. Pairs of samples are then drawn from Gaussians $i$ and $j$ with a fixed diagonal covariance matrix $\tau I$. When sampling matching pairs $i=j$; for non-matching pairs $i\neq j$. $10,000$ training samples and $10,000$ test samples are used, half of which are matching and half non-matching.

The representation networks had three hidden layers of size $256$ and ReLU activations. Four Siamese networks were trained, based on the four DrLIM variants: SPRING, CENTRIFUGE, SPRING+SD, and CENTRIFUGE+SD.

\begin{figure*}[!t]
\begin{minipage}[c]{0.60\textwidth}
\begin{tabular}{cc}
\includegraphics[trim=55 35 50 30,clip,width=.45\linewidth]{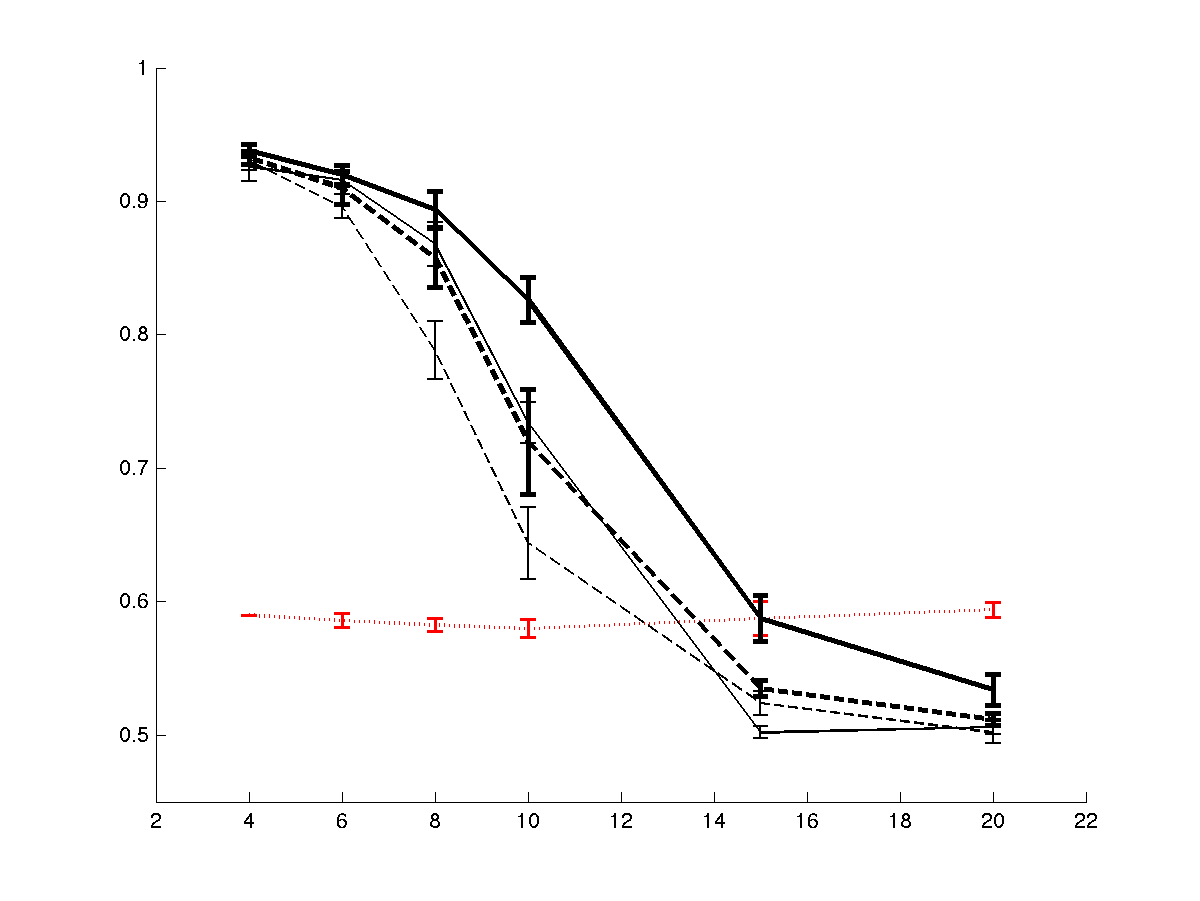}&
\includegraphics[trim=55 35 49 30,clip,width=.45\linewidth]{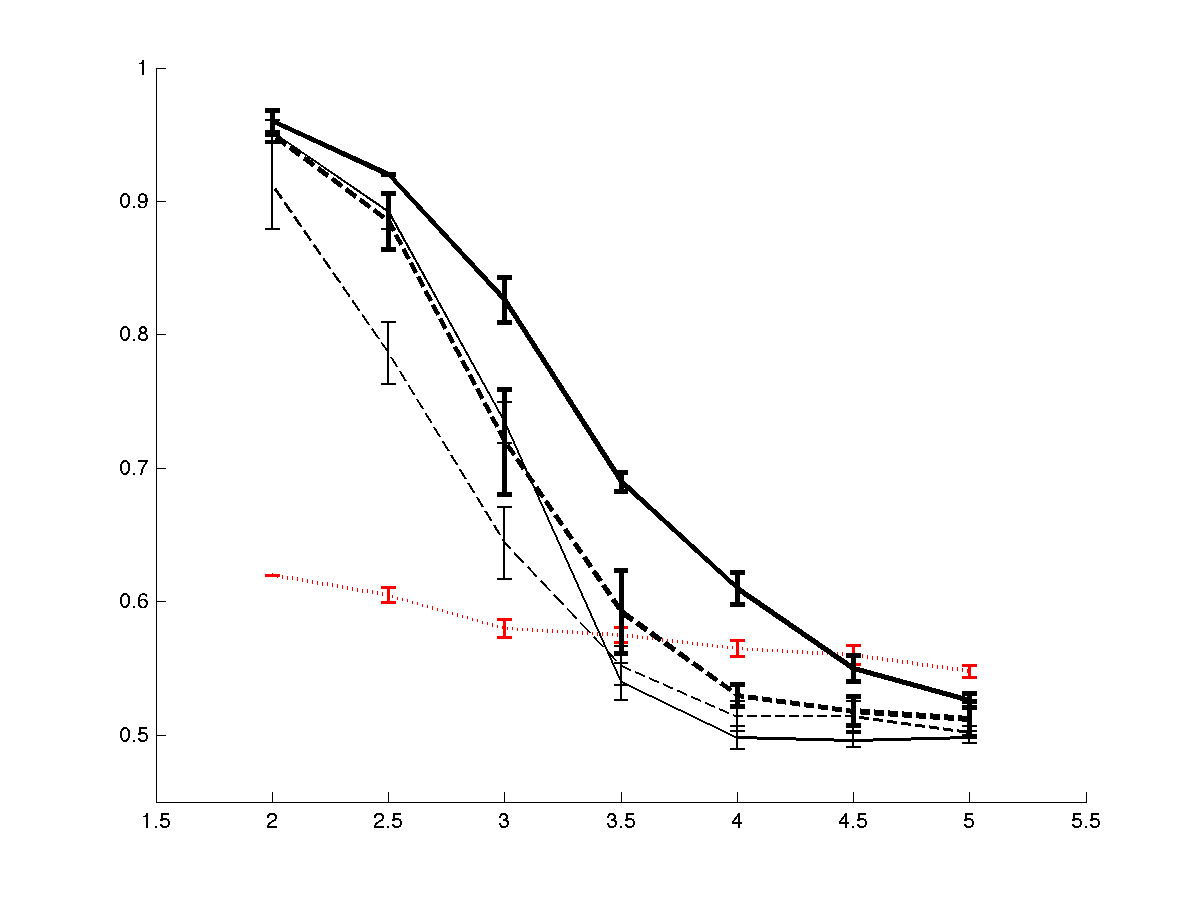}\\
(a) & (b)\\
\includegraphics[trim=55 35 50 30,clip,width=.45\linewidth]{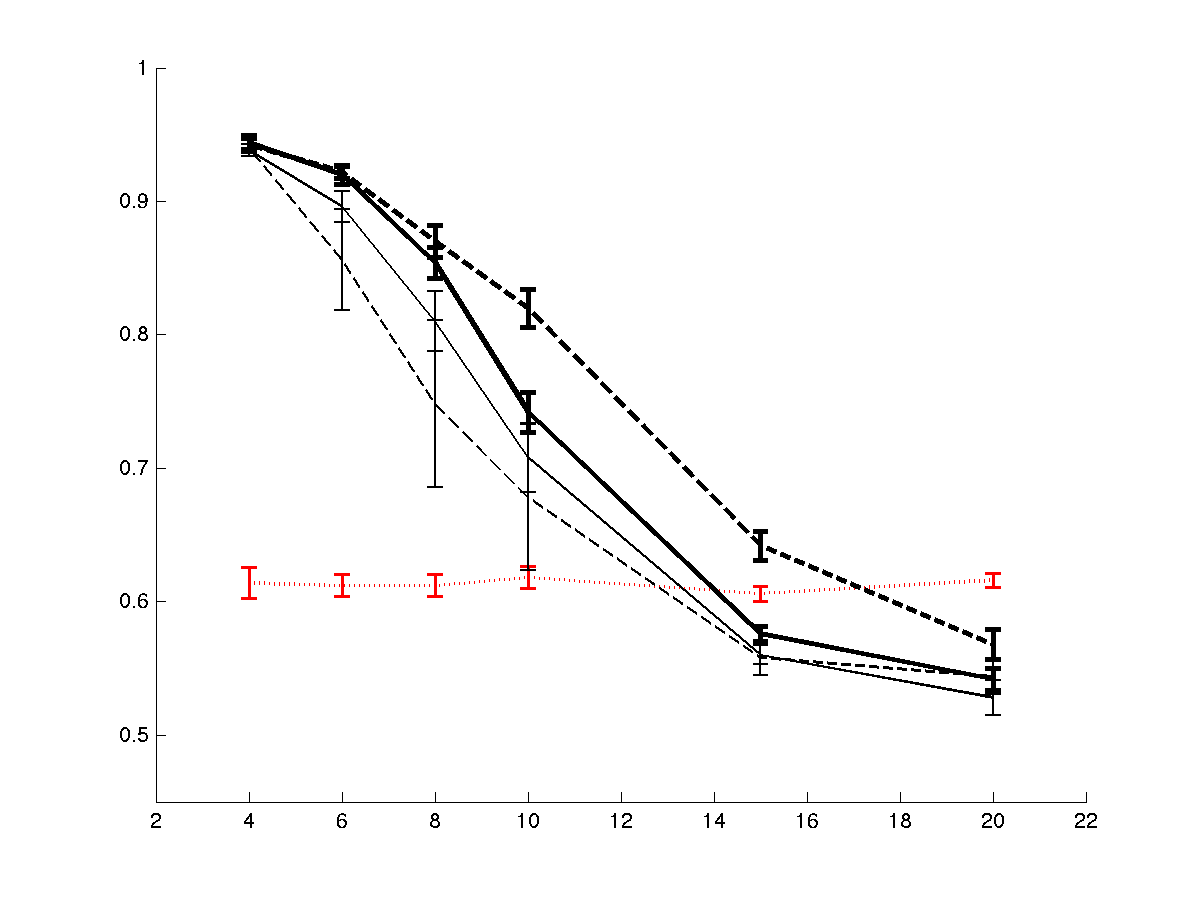}&
\includegraphics[trim=55 35 49 30,clip,width=.45\linewidth]{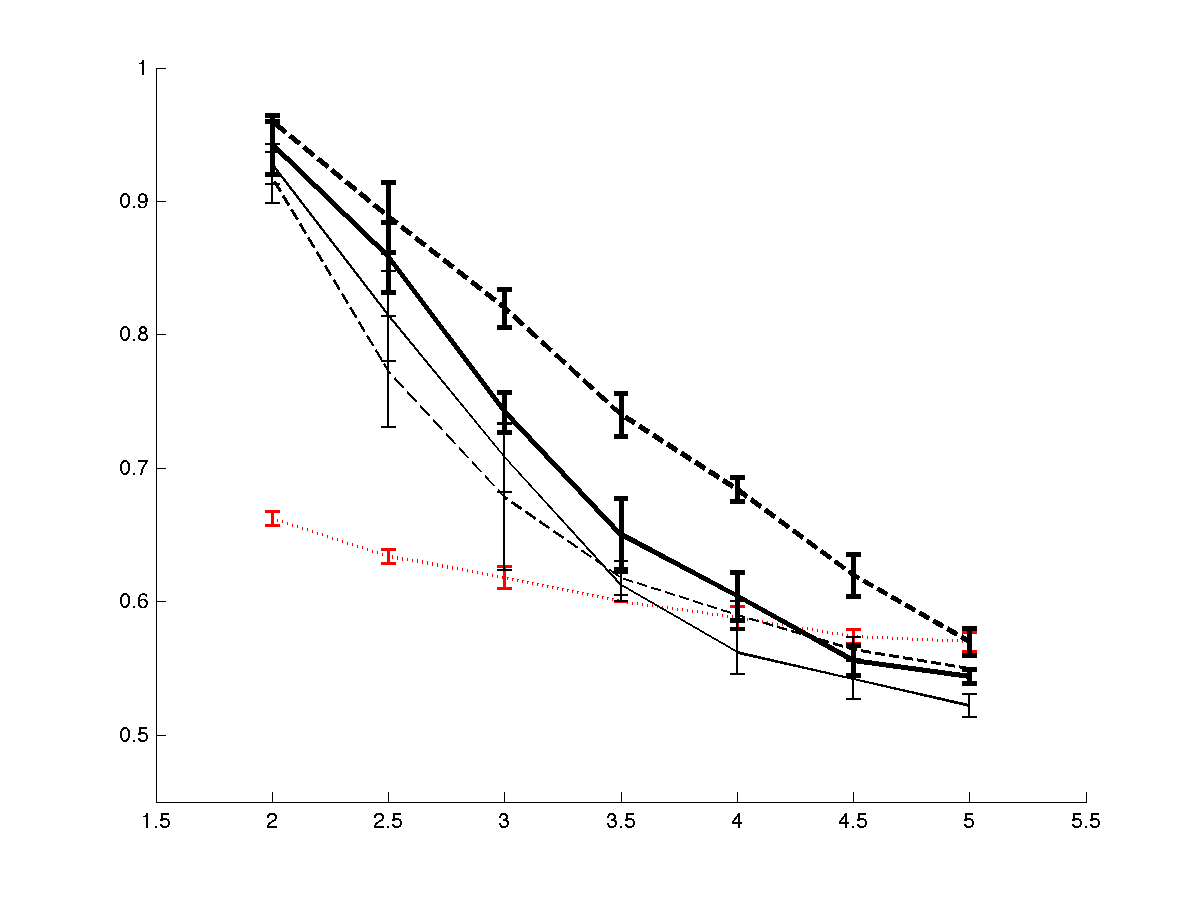}\\
(c) & (d)\\
\end{tabular}
\end{minipage}\hfill
\begin{minipage}[c]{0.4\textwidth}
\caption{Results of the synthetic experiment. The first row shows the results of the baseline experiment. The second row shows the results where each datapoint was normalized to have a fixed norm. All plots show mean and SD of AUC (y-axis) obtained for five types of features. In all plots, the faint (red) dotted line presents the original random features sampled, as describe in Sec.~\ref{sec:exp}. The thin solid line presents the results obtained for the original DrLIM method (SPRING). The thin dashed line shows the results for the CENTRIFUGE variant. The thick solid and dashed lines show the respective counterparts where SD was added to the loss. (a) and (c) present results when varying (x-axis) the number of Gaussians $n_c$ from which points where sampled. (b) and (d) explore the effect of changing the variance parameter $\tau$. As can be seen, SD improves performance in almost all experiments. For the baseline data SPRING outperforms CENTRIFUGE. The situation is reversed when the norm of the sampled datapoints is fixed.}
\label{fig:syn}
\end{minipage}
\end{figure*}

Two sets of experiments are conducted. In the first set, $\tau=3$ and $n_c$ varies between 4 and 20. In the second set $n_c=10$ and $\tau$ varies between 2 and 5. Each setting is repeated 10 times, and the plots in Fig.~\ref{fig:syn}(a),(b) depict the mean Area Under Curve (AUC) obtained when training the network on the training data and evaluating on the test data for the first and the second set respectively. As can be seen, in almost all experiments, SPRING outperforms CENTRIFUGE and SPRING+SD outperforms CENTRIFUGE+SD. It is also clear that the SD versions of each physical model greatly outperform the vanilla versions.

The entire experiment was then repeated, with a slight variant. In the second variant, the sampling process is identical except that the two samples in each pair are both normalized to have a norm of one. The exact same experiments were repeated. In Fig.~\ref{fig:syn}(c), $n_c$ varies while $\tau=3$ is fixed. In Fig.~\ref{fig:syn}(d), $\tau$ varies while $n_c=10$. In these experiments, the SD version also outperforms the plain SPRING and CENTRIFUGRE versions by a large margin. However, among the physical models the leading performance for the normalized inputs is obtained using the CENTRIFUGE method. This is true for both the SD and the vanilla variants.

In all experiments performed we have added a baseline method, which is the norm of the difference between the pairs of points. This method does moderately better than chance (AUC of about 0.6) and is, in general, much inferior to the network representations. However, when the number of classes dramatically increases, or when the variance is very high, this simple method has an advantage over the learned models.

One additional experiment we conducted is to evaluate our method using the "accuracy@10" measure proposed in \cite{deepmatching}. "Accuracy@10" is defined as the proportion of correct assignments from the first image to the second with respect to the total number of pixels. A pixel assignment is considered correct if its Euclidean error is smaller than 10 pixels. The "accuracy@10" score achieved by DeepMatching~\cite{deepmatching} on the KITTI2012 dataset is 0.856. We computed the same score using our descriptors on our own validation set (last 20\% of images by file order) and achieved a score of 0.960.

\label{sec:exp}


\subsection{Comparison of loss variants on optical flow datasets}

In our optical flow experiment, we make use of the three largest and most competitive datasets: KITTI2012 \& KITTI2015, both of which contain real image datasets taken from a moving vehicle in a city environment and MPI-Sintel, which is an extensive computer graphics dataset. 

We ran the four variants up to 1500 epochs while conducting the comparison. The margin parameter $m$ was determined, for each variant, using initial runs of 500 epochs. The performance was evaluated on a set of images set aside for this purpose: 20\% of the images of the KITTI2012 and KITTI2015 training sets, which come last in the file order, and a random sample of 50 images of the FINAL training subset of MPI-Sintel.

The results are reported in Tab.~\ref{tab:1500_12} and ~\ref{tab:1500_ms} for KITTI2012, KITTI2015, and MPI-Sintel respectively. Each table compares the four variants: SPRING, CENTRIFUGE, SPRING+SD, and CENTRIFUGE+SD. The nature of the error rate used depends on the dataset conventions: in KITTI2012 and KITTI2015, the percent of pixels that displaced more than 3 pixels (Euclidean error) from the ground truth is used; in MPI-Sintel, the mean end point error is reported for all and matching-only pixels. In the KITTI2012 and KITTI2015 lines two error rates are reported in each cell: one obtained after the PM matching process only, and the second error after applying the interpolation process. 

One can observe a consistent drop in the error rate when shifting from the SPRING model to the CENTRIFUGE model, especially prior to the interpolation. There is an additional consistent drop in error when adding an SD term to either losses. Based on these partial experiments, we decided to focus on the CENTRIFUGE+SD method and train using this variant for 4,000 epochs on each of the datasets. 


\begin{table}[t]
\begin{center}
\begin{tabular}{|l|c|c|c|}
\hline
Loss & Epoch 500 & Epoch 1000 & Epoch 1500 \\
\hline\hline
{KITTI'12:} & & &\\
SPRG & 10.48 / 5.13 & 10.11 / 5.05 & 10.04 / 4.96 \\
CENT & 9.93 / 5.19 & 9.57 / 4.91 & 9.54 / 4.76 \\
SPRG+SD & 9.11 / 5.03 & 8.97 / 4.92 & 8.64 / 4.88 \\
CENT+SD & 8.91 / 4.85 & 8.99 / 4.95 & 8.54 / 4.97 \\
\hline
KITTI'15: & & &\\
SPRG & 29.7 / 19.97 & 29.98 / 19.49 & 28.74 / 19.43 \\
CENT& 29.8 / 20.59 & 28.24 / 19.4 & 27.92 / 18.62 \\
SPRG+SD & 27.41 / 19.30 & 26.29 / 18.91 & 27.00 / 18.95 \\
CENT+SD & 28.20 / 20.40 & 27.02 / 19.19 & 26.34 / 19.05 \\
\hline
\end{tabular}
\end{center}
\caption{Loss comparison on KITTI2012 and KITTI2015 after a certain number of epochs. Each row is a different variant of DrLIM, see Section~\ref{sec:loss}. Each cell shows the \% of pixels with euclidean error $>$ 3 pixels after the ANN process (left) and after bidirectional consistency check and EpicFlow interpolation (right).}
\label{tab:1500_12}
\end{table}

\begin{table}[t]
\begin{center}
\begin{tabular}{|l|c|c|c|}
\hline
Loss & Epoch 500 & Epoch 1000 & Epoch 1500 \\
\hline\hline
SPRG & 3.17 / 2.40 & 3.27 / 2.54 & 3.06 / 2.39 \\
\hline
CENT & 3.44 / 2.59 & 3.34 / 2.60 & 3.41 / 2.63 \\
\hline
SPRG+SD & 3.42 / 2.63 & 3.43 / 2.53 & 3.05 / 2.17 \\
\hline
CENT+SD & 3.25 / 2.49 & 3.15 / 2.32 & 3.15 / 2.36 \\
\hline
\end{tabular}
\end{center}
\caption{Comparing DrLIM variants on MPI-Sintel. Presented results are post EpicFlow interpolation. Shown are average EPE (end-point-error) on all the pixels in the images (left) and EPE on valid pixels (as defined by the dataset) (right) on the FINAL pass.}
\label{tab:1500_ms}
\end{table}

\subsection{Benchmark results}

We trained our main architecture ($51\times51$ patch-size, CENTRIFUGE+SD loss) on all three datasets. The network architecture is identical in all three cases. As a training set for KITTI2012 and KITTI2015, we took the first 80\% of the image pairs and as a validation set, the remaining 20\%. For MPI-Sintel we chose 80\% of the image pairs for training and the rest for validation. We chose 2M random samples out of those 20\% images to act as the validation samples during training. Training was performed for 4000 epochs, and the configuration with the best validation loss was recorded and deployed. 

As can be seen in Tab.~\ref{tab:k2012},~\ref{tab:k2015},~\ref{tab:ms}, we were able to achieve state-of-the-art results on the official KITTI2012 and KITTI2015 benchmarks, and rank in the 6th place on the MPI-Sintel benchmark. The gap in ranking between the KITTI datasets and MPI-Sintel might arise from the fact that we are the only top reported system that does not use color on MPI-Sintel. 

Since CENTRIFUGE+SD was not clearly preferable on MPI-Sintel to other methods by epoch 1500 (Tab.~\ref{tab:1500_ms}), we submitted results for all 4 DrLIM variants on this benchmark. The obtained order of results (Tab.~\ref{tab:ms}) is CENTRIFUGE+SD, SPRING, SPRING+SD, and CENTRIFUGE. A significant gap of 0.4 EPE exists between CENTRIFUGE+SD and SPRING.

On KITTI2012, we have also submitted the predictions of the FAST network, in which our fine-grained batch normalization (Sec.~\ref{sec:networkarchitecture}) is replaced with the conventional batch normalization. There are only four methods that are ranked between the ACCURATE and the FAST methods.

\begin{table}[t]
\begin{center}
\begin{tabular}{|l|c|c|}
\hline
Method & Out-Noc & Running time\\
\hline\hline
{\bf PatchBatch-ACCRTE-PS71} & { 5.29\%} & 60.5s\\
{\bf PatchBatch-ACCURATE} & { 5.44\%} & 50.5s\\
PH-Flow~\cite{phflow} & 5.76\% & 800s\\
FlowFields~\cite{flowfields} & 5.77\% & 23s\\
CPM-Flow (anon.) & 5.80\% & 2s\\
NLTGV-SC~\cite{NLTGV-SC} & 5.93\% & 16s\\
{\bf PatchBatch-FAST} & 5.94\% & 25.5s \\ 
DDS-DF~\cite{DDS-DF} & 6.03\% & 1m \\
TGV2ADCSIFT~\cite{TGV2ADCSIFT} & 6.20\% & 12s \\
DiscreteFlow~\cite{discreteflow} & 6.23\% & 3m \\
\hline
\end{tabular}
\end{center}
\caption{Top 10 KITTI2012 2-frame (Pure) Optic Flow Algorithms as published on the submission date. Out-Noc is the percentage of pixels with euclidean error $>$ 3 pixels out of the non-occluded pixels}
\label{tab:k2012}
\end{table}

\begin{table}[t]
\begin{center}
\begin{tabular}{|l|c|c|}
\hline
Method & Fl-all & Running time\\
\hline\hline
{\bf PatchBatch-ACCURATE} & 21.69\% & 50.5s\\
DiscreteFlow~\cite{discreteflow} & 22.38\% & 3min\\
CPM-Flow (anon.) & 24.24\% & 2s\\
EpicFlow~\cite{epicflow} & 27.10\% & 15s\\
FilteringFlow (anon.) & 28.50\% & 116s\\
DeepFlow~\cite{deepflow} & 29.18\% & 17s \\
HS~\cite{HS} & 42.18\% & 2.6m \\
DB-TV-L1~\cite{DB-TV-L1} & 47.97\% & 16s \\
HAOF~\cite{HAOF} & 50.29\% & 16.2s\\
PolyExpand~\cite{polyexpand} & 53.32\% & 1s\\
\hline
\end{tabular}
\end{center}
\caption{Top 10 KITTI2015 2-frame Optic Flow Algorithms as of the submission date. Fl-all is the percentage of pixels with euclidean error $>$ 3 pixels. The FAST network was not trained on this benchmark by the submission time.}
\label{tab:k2015}
\end{table}

\begin{table}[t]
\begin{center}
\begin{tabular}{|l|c|c|}
\hline
Method & EPE all, `final' pass \\
\hline\hline
FlowFields~\cite{flowfields} & 5.810 \\
CPM-Flow (anon.) & 5.960 \\
DiscreteFlow~\cite{discreteflow} & 6.077 \\
EpicFlow~\cite{epicflow} & 6.285 \\
Deep+R~\cite{deep+r} & 6.769 \\
{\bf PatchBatch-CENT+SD} & 6.783 \\
DeepFlow2 (anon.) & 6.928 \\
{\bf PatchBatch-SPRG} & 7.188 \\
SparseFlowFused~\cite{sparseflowfused} & 7.189 \\
DeepFlow~\cite{deepflow} & 7.212 \\
FlowNetS+ft+v~\cite{flownet} & 7.218 \\
NNF-Local~\cite{nnflocal} & 7.249 \\
{\bf PatchBatch-SPRG+SD} & 7.281 \\
{\bf PatchBatch-CENT} & 7.323 \\
SPM-BP~\cite{spmbp} & 7.325 \\
AggregFlow~\cite{aggregflow} & 7.329\\
\hline
\end{tabular}
\end{center}
\caption{Top MPI-Sintel results as of the submission date. Each number represents the EPE (end-point-error), averaged over all the pixels in the comparison images, using the 'final' rendering pass of MPI-Sintel. Four ACCURATE variants are shown. The CENTFIGURE+SD network is ranked 6th as of the paper's submission date. The TF+OFM method~\cite{TF+OFM} (EPE 6.727) is removed from this table since it is not a pure 2-frame optical flow method.}
\label{tab:ms}
\end{table}

\subsection{Network variants}

We explored several network variants on the KITTI2012 validation benchmark. These variants explore different descriptor sizes and different patch sizes, in addition to our variant of batch normalization. 

The results of these experiments are displayed in Tab.~\ref{tab:variants}. The table shows the percentage of pixels with displacement error larger than 3 pixels after the ANN matching process and after the interpolation process. The full (``ACCURATE'') method is compared with the FAST network. We also compared to an ACCURATE network in which the input patch size is $71 \times 71$ pixels. Two other variants in which the final descriptor size varies are shown. The descriptor size was altered by replacing Conv5's filter-size to $1 \times 1$ to obtain a 1024D descriptor, or by adding an additional convolutional (and batch-normalization) layer with 32 feature maps to obtain a 32D descriptor. The 1024D descriptor makes PM run much slower. The converse is true for 32D. Based on these results, further improvements of our method's accuracy are expected with larger patch and representation sizes. 

In another experiment we tested the ACCURATE network trained on KITTI2015 on the KITTI2012 validation images. The performance seems comparable to that of the KITTI2012 ACCURATE network, attesting to the generality of the learned patch matching function.

Our method was designed with the requirement of obtaining a generic pipeline that employs L2 distances of patches. In this way, the ANN and interpolation methods can be replaced with other, perhaps more efficient methods, and the gain in performance can be preserved. The  running time of each step of the computation for the baseline and the FAST methods are detailed in Tab.~\ref{tab:runtime}. The patch encoding process is the only process currently done on the GPU. Its running time dominates the ACCURATE network's execution time, but is less than 10\% of that of the FAST network.

\begin{table}[t]
\begin{center}
\begin{tabular}{|l|c|c|}
\hline
Method & KITTI2012 err& Encode time \\
\hline\hline
ACCURATE & 8.08 / 4.80 & 27s\\
FAST & 9.45 / 5.3 & 2.5s\\
ACCURATE $71 \times 71$ & 7.85 / 4.79 & 37s\\
ACCURATE 32D & 9.34 / 5.23 & 27s \\
ACCUARTE 1024D & 8.10 / 4.81 & 37s \\
\hline
Train on KITTI2015 & 8.99 / 4.97 & 27s\\
\hline
\end{tabular}
\end{center}
\caption{Additional variants comparison on KITTI2012. All results are reported using the CENTRIFUGE+SD loss, while taking the model with the lowest loss on validation data out of 4000 epochs. The error is computed on the local validation set. Each row presents the \% of pixels with euclidean error $>$ 3 pixels after the ANN process (left) and after the interpolation and bidirectional consistency check (right). In addition, the time it takes to compute the patch descriptors in seconds is shown. As can be seen, additional improvement for our method is expected when using larger patches and a longer representation vector.}
\label{tab:variants}
\end{table}

\begin{table}[t]
\begin{center}
\begin{tabular}{|l|c|c|}
\hline
Step & ACCURATE & FAST \\
\hline\hline
Descriptor computation& 27s & 2.5s\\
ANN (PatchMatch) & 6.5s& 6.5s\\
Connected component analysis & 0.5s & 0.5s \\ 
Interpolation (EpicFlow) & 16s & 16s \\
\hline
Total & 50s & 25.5s \\
\hline
\end{tabular}
\end{center}
\caption{The runtime of our ACCURATE network (using fine-grained batch normalizatoin) and the FAST method. The descriptor computation is done in parallel for the two images and so are  the PatchMatch computations per direction.}
\label{tab:runtime}
\end{table}

\section{Discussion and future work}

Using CNNs for encoding each patch separately leads to a solution that is entirely flexible. On one hand the CNN can be modified, pruned, or compressed~\cite{fitnet} in order to control the accuracy to run time trade-off. On the other hand, the other steps can be replaced, implemented on the GPU, or bypassed as needed. A fast alternative, for example, for the ANN solution employed is the kd-tree solution of~\cite{kdtree}. Our reliance on simple vector representations means that this integration does not require any modification.

The problem of metric learning is a central Machine Learning task that is used in computer vision domains ranging from low-level vision to almost all high level vision tasks. Mahalanobis distances, and other distances that translate to $L2$ matching of learned representations dominate the relevant literature. 

The DrLIM loss is a prominent solution for learning L2 distances using deep networks. We believe that the two orthogonal types of improvements that we presented here can lead not only to state of the art optical flow, but also to improved results in many other domains. The success on what might be the simplest imaginable (and therefore the most general) synthetic data is highly suggestive of that.

In addition to this very general contribution, the very idea of using batch losses is novel, as far as we know. Losses are always constructed per sample and then aggregated. This locality is compatible with the stochastic gradient descent. However, when using mini batches, per batch losses are also compatible.

Batch losses can tie together the samples in a batch and support the design of networks that take into account interrelations between the samples in the batch. We have demonstrated the effectiveness of this approach in the domain of metric learning. Future work might take advantage of this in order to whiten the representation layer, whiten the error of  regressors along the output dimensions, or balance the error between the classes in a multiclass scenario. 

\section*{Acknowledgments}
This research is supported by the Intel Collaborative Research Institute for Computational Intelligence (ICRI-CI). The authors would like to thank Michael Rotman for valuable insights.

%

\clearpage
{\small
\bibliographystyle{ieee}
\bibliography{patchflow}
}

\end{document}